\documentclass{article}

\usepackage{arxiv}

\usepackage{cite}

\newcommand{\citep}[1]{\cite{#1}}
\newcommand{\citet}[1]{\cite{#1}}

\usepackage{amsmath,amssymb,amsthm}
\usepackage{dsfont,mathtools}
\usepackage{xspace}
\usepackage{epsfig}
\usepackage{algorithm}
\usepackage[noend]{algorithmic}
\usepackage{color}
\usepackage{url}
\usepackage{graphicx}
\usepackage{tikz}
\usepackage{float}
\usepackage{booktabs}
\usepackage{makecell}
\usepackage{multicol}
\usepackage{multirow}
\usepackage{csquotes}
\usepackage{wasysym}
\usepackage{longtable}
\usepackage{colortbl}
\usepackage{paralist}
\usepackage{hyperref}
\usepackage{ragged2e}

\definecolor{colorEAXR}{RGB}{27,147,108}
\definecolor{colorEAXRGPX}{RGB}{212,85,4}
\definecolor{colorLKHRIPT}{RGB}{105,100,170}

\definecolor{ercisred}{RGB}{133, 35, 57}
\definecolor{ercisblue}{RGB}{67, 92, 139}

\hypersetup{
    colorlinks = true,
    linkcolor  = ercisblue,
    urlcolor   = ercisred,
    citecolor  = ercisblue,
}

\title{To Boldly Show What No One Has Seen Before: A Dashboard for Visualizing Multi-objective Landscapes}

\author{
  Lennart Sch{\"a}permeier\\
  Statistics and Optimization\\
  University of M{\"u}nster \\
  M{\"u}nster, Germany \\
  \texttt{schaepermeier@uni-muenster.de} \\
   \And
  Christian Grimme \\
  Statistics and Optimization\\
  University of M{\"u}nster \\
  M{\"u}nster, Germany \\
  \texttt{christian.grimme@uni-muenster.de}
  \And
  Pascal Kerschke \\
  Statistics and Optimization\\
  University of M{\"u}nster \\
  M{\"u}nster, Germany \\
  \texttt{kerschke@uni-muenster.de}
}

\begin{document}
\maketitle

\pagestyle{plain}
\thispagestyle{fancy}
\lfoot{\vspace*{-0.5cm}\rule{\columnwidth}{0.8pt}\\
\footnotesize \begin{justify}This version has been accepted for publication at the \textit{11th International Conference on Evolutionary Multi-Criterion Optimization March 28-31, 2021, in Shenzhen, China}. Permission from the authors must be obtained for all other uses, in any current or future media, including re\-printing/re\-pub\-lishing this material for advertising or promotional purposes, creating new collective works, for resale or redistribution to servers or lists, or reuse of any copyrighted component of this work in other works.\end{justify}}\cfoot{}

\begin{abstract}
Simultaneously visualizing the decision and objective space of continuous multi-objective optimization problems (MOPs) recently provided key contributions in understanding the structure of their landscapes. For the sake of advancing these recent findings, we compiled all state-of-the-art visualization methods in a single R-package (\texttt{moPLOT}). Moreover, we extended these techniques to handle three-dimensional decision spaces and propose two solutions for visualizing the resulting volume of data points. This enables -- for the first time -- to illustrate the landscape structures of three-dimensional MOPs.

However, creating these visualizations using the aforementioned framework still lays behind a high barrier of entry for many people as it requires basic skills in R. To enable any user to create and explore MOP landscapes using \texttt{moPLOT}, we additionally provide a dashboard that allows to compute the state-of-the-art visualizations for a wide variety of common benchmark functions through an interactive (web-based) user interface.

\end{abstract}

\section{Introduction} \label{sec:introduction}

When analyzing and developing optimization methods, scientists and practitioners from Operations Research traditionally use methods for visualizing problems and their characteristics. But also teaching Operations Research without graphical representation of search and objective space (often together in a functional landscape) is unthinkable: no textbook of Operations Research can do without the classical visual presentation of problem characteristics, showing multimodality, minima, maxima, plateaus, etc.~\cite{preuss2015}.

For most of the last two decades, there has been only one approach in (continuous) multi-objective optimization~\cite{Fonseca1995}, which provided analogous representations to the single-objective domain. Apart from the representation of the Pareto front, and in a few cases the Pareto set~\cite{CC07,miettinen2012nonlinear}, only few researchers invested significant effort into the advancement of visualization \cite{Tusar15tevc,Filipic2020}. Accordingly, the methodology for visualization in the domain of multi-objective optimization is still in its infancy. Yet, recently a few more methods have been introduced~\cite{kerschke2017expedition,schapermeier2020plot}, which help to study multi-objective problems from a new perspective.

In fact, the recent insights into the decision space of multi-objective optimization problems (MOPs) have contributed greatly to revealing structural characteristics and to the understanding of benchmark problem landscapes~\cite{schapermeier2020plot} and contributed to the development of algorithmic ideas. That lead to an enhanced multi-objective gradient descent, which follows local efficient sets until sliding into dominating basins of attraction, eventually even leading to the Pareto set in many cases~\cite{GrimmeKEPDT2019SlidingToThe,GrimmeKT2019Multimodality}. Further, these insights helped in the development of advanced single-objective local search, which is capable of escaping local optima by multiobjectivization and exploiting the known structures~\cite{steinhoff_multiobjectivization_2020,aspar2021multi3}.

In order to provide the new visualization techniques for researchers and practitioners dealing with MO problems
an \texttt{R}-package, \texttt{moPLOT}, has been published. However, this package is not accessible for anyone without knowledge of the scripting language \texttt{R}. This makes it difficult to analyze even well-known benchmarks. In addition, the current visualization techniques are often limited to bi-objective problems and two-dimensional decision spaces. Although visualization is naturally limited to three dimensions, the current methods do not exploit this potential w.r.t. search space. In contrast to common single-objective landscapes, solution quality (in the MO context expressed by mutual dominance or incomparability of solutions) is not visualized as height but only colored accordingly. In principle, that leaves room for considering visualization of a third dimension in decision space. Of course, this implies that a volume of solutions is shown, with the inner solutions being hidden by solutions further out. Accordingly, a  technique for visually accessing the inner solutions has to be provided. Addressing all aforementioned issues, the key contributions of this work are the following:
\begin{enumerate}
\item We extend the R-package \texttt{moPLOT} by an interactive and user-friendly (web-based) dashboard, which allows users with little programming experience to visually explore the problem landscapes of common continuous MO benchmarks with two and three objectives and in two- and three-dimensional decision space settings.
\item Enabled by the interactive dashboard environment, we (a) extend the existing methods to cope with up to three dimensions, and (b) propose two views on the resulting volume of points, paving the way for visualizing MOPs with up to three variables and/or objectives.
\end{enumerate}
While the first contribution enables many researchers to peek into before unseen (and for many still surprising) structures of MOPs, the latter allows to deepen the understanding of these structures and provides some generalization of the currently 2D-based interpretation of MO landscapes.

The remainder of this paper first provides some background on continuous multi-objective optimization and the state-of-the-art techniques for visualizing MOPs, followed by an introduction to our dashboard application and its core features. In section \ref{sec:three-dimensional}, we then present our extensions of the existing visualizations to three-dimensional decision and objective spaces. Finally, section \ref{sec:conclusions} gives some concluding remarks and an outlook on future research potentials.

\section{Background} \label{sec:background}

In the following, we will briefly introduce important concepts and methods that are underlying this work. While section~\ref{sec:moo} introduces the mathematical terminology used within this work, section~\ref{sec:visualizations} will give an overview of the state-of-the-art methods for illustrating the landscapes of continuous MOPs.

\subsection{Continuous Multi-objective Optimization} \label{sec:moo}

We consider box-constrained continuous MOPs $f: \mathbb{R}^p \rightarrow \mathbb{R}^k$ with $p$ decision variables and $k$ objectives, which are w.l.o.g.~to be minimized. The box constraints limit the decision space to a feasible domain of $\mathcal{X} := [\mathbf{l}, \mathbf{u}] \subset \mathbb{R}^p$:
\begin{align}
    \begin{split}
        f(\mathbf{x}) = (f_1(\mathbf{x}), \dots, f_k(\mathbf{x})) \overset{!}{=} \min \qquad \text{ with } \qquad l_i \leq x_i \leq u_i, \; i=1, \dots, p.
    \end{split}
    \label{eq:mop}
\end{align}
For the remainder of this work, we consider two- and three-dimensional decision and objective spaces, i.e., $p \in \{2, 3\}$ and $k \in \{2, 3\}$.

In general, the solution of a MOP is the Pareto set $\mathcal{X}^* \subseteq \mathcal{X}$ of non-dominated solutions, i.e., the set of solutions $\mathbf{x}^* \in \mathcal{X}$ for which there exists no solution $\mathbf{x} \in \mathcal{X}$ with $f_i(\mathbf{x}) \leq f_i(\mathbf{x}^*)$, $i = 1, \ldots, k$, and $f_i(\mathbf{x}) < f_i(\mathbf{x}^*)$ for at least one $i$. The image of the Pareto set under $f$ is known as the Pareto front $f(\mathcal{X}^*)$.

Similar to single-objective continuous optimization, local optimality in the multi-objective case is also defined relative to an $\varepsilon$-ball $B_\varepsilon(\mathbf{x}) \subseteq \mathcal{X}$~\cite{GrimmeKEPDT2019SlidingToThe}. That is, a solution $\mathbf{x}$ is considered locally optimal (i.e., it is a locally efficient point), if it is not dominated by any other solution $\mathbf{y} \in B_\varepsilon(\mathbf{x})$ for some $\varepsilon$.

A useful concept when studying local optimality, and an important tool for some of the visualizations, is the multi-objective gradient (MOG)\footnote{Note that we focus here on the case where the box constraints are inactive. Some discussion of how to handle the decision boundary can be found, e.g., in \cite{schapermeier2020plot}.}. It points towards a common descent direction for all objectives, if one exists, and vanishes if a locally efficient point is reached. In the bi-objective case, a multi-objective gradient is found by the sum of the normalized single-objective gradients \cite{kerschke2017expedition}. This definition, however, does not generalize for $k > 2$.

A more general definition of a MOG is given by the shortest vector in the convex hull of the normalized single-objective gradients \cite{desideri2012multiple}. In the bi-objective case, this can be simplified to almost yield the previous definition, up to an additional factor of $\frac 1 2$:
\[
\nabla f(\mathbf{x}) := (\mathbf{u}_1 + \mathbf{u}_2) / 2 \qquad \text{ where } \qquad \mathbf{u}_i := \nabla f_i(\mathbf{x}) / \|\nabla f_i(\mathbf{x})\| \text{ with } i = 1, 2
\]
if all single-objective gradients are non-zero and $\nabla f(\mathbf{x}) := 0$ otherwise.

\begin{figure}[t!]
    \centering
    \includegraphics[height=3.7cm]{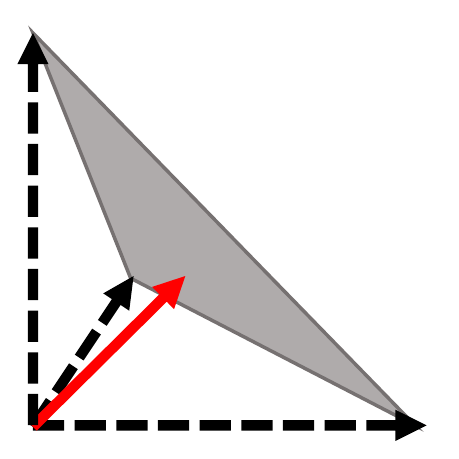}\qquad
    \includegraphics[height=3.7cm]{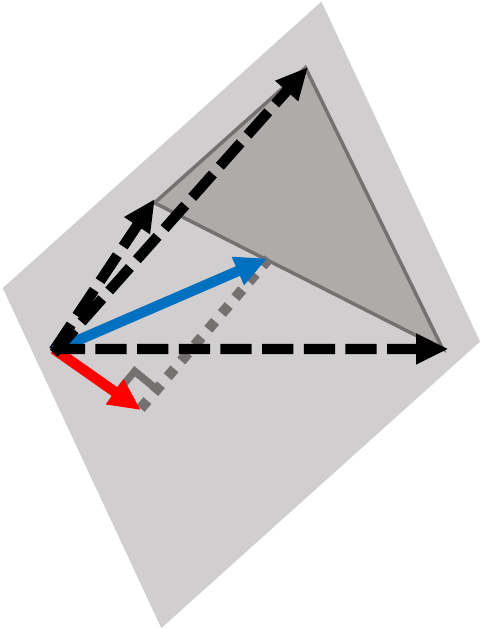}
    \vspace*{-0.1cm}
    \caption{Main cases for the tri-objective MOG in 3D. The (normalized) single-objective gradients are dashed. The shortest vector to the plane spanned by the three gradients is shown in red. If it lies inside their convex hull, this is the MOG (left image). Otherwise (right image), if it lies outside the convex hull, the MOG is given by the shortest bi-objective gradient (blue).}
    \vspace*{-0.4cm}
    \label{fig:tri_obj_mog}
\end{figure}

For the tri-objective scenario ($k = 3$), we distinguish two cases. (1) For $p = 2$ decision variables: If the convex hull of the single-objective gradients does not contain the zero vector, the MOG is given by the bi-objective MOG of the two single-objective gradients with the largest angle between them, otherwise it is defined as $0$. (2) For $p = 3$ decision variables, we can compute the orthogonal projection of the origin onto the plane that is spanned by the three normalized single-objective gradients. If the projection is located within the gradients' convex hull, this gives the MOG. Otherwise, the MOG touches the boundary of the convex hull, and it can be found as the shortest of the \textit{pairwise} bi-objective gradients. A schematic illustration of this case is given in Figure \ref{fig:tri_obj_mog}.

\subsection{Visualizations of Continuous Multi-objective Landscapes}
\label{sec:visualizations}

\begin{figure}[t!]
    \centering
    \includegraphics[width=0.32\textwidth]{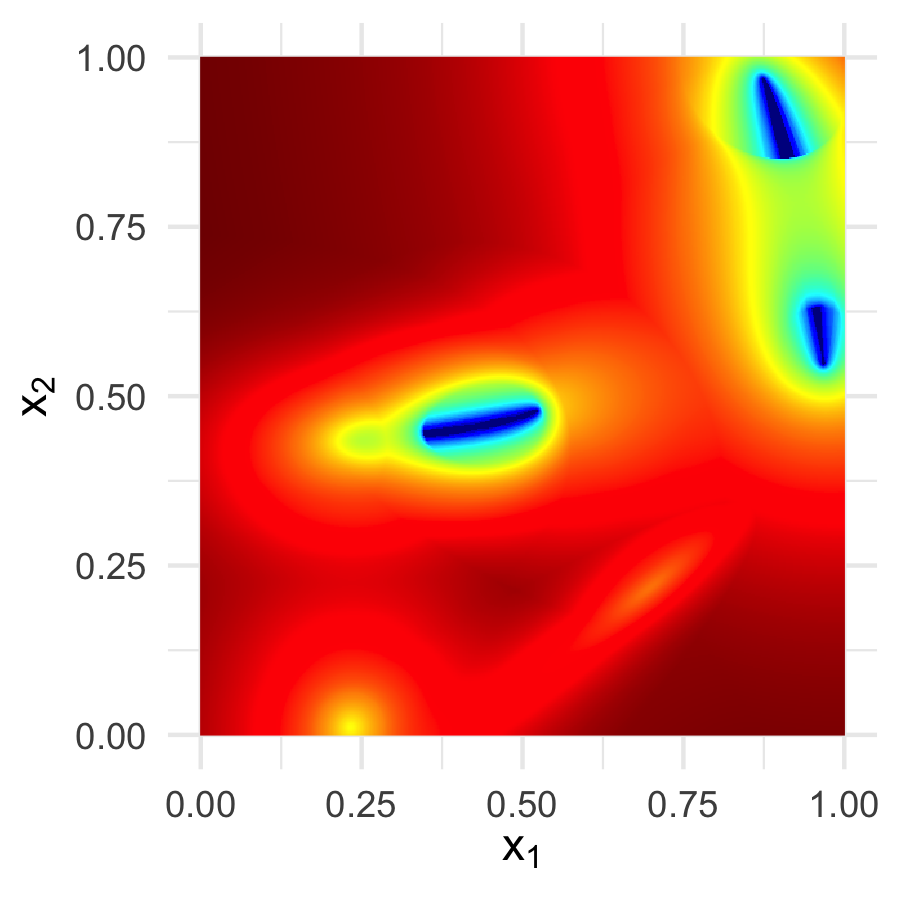}
    \hfill
    \includegraphics[width=0.32\textwidth]{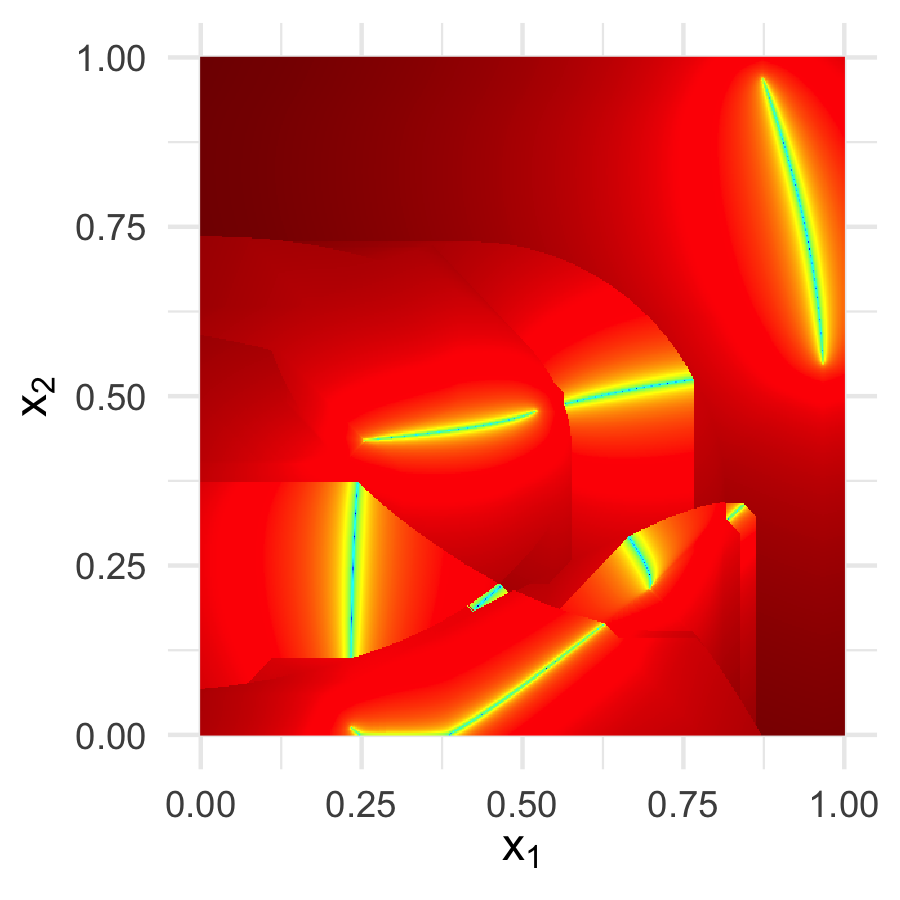}
    \hfill
    \includegraphics[width=0.32\textwidth]{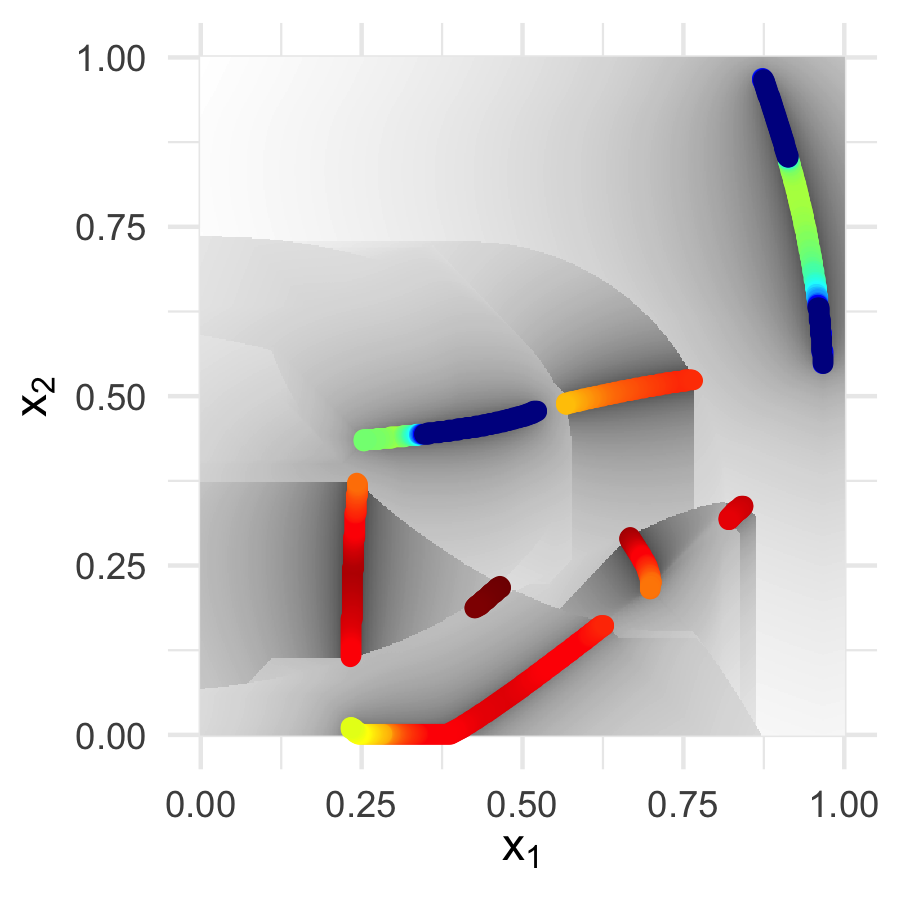}
    \hfill
    \includegraphics[width=0.32\textwidth]{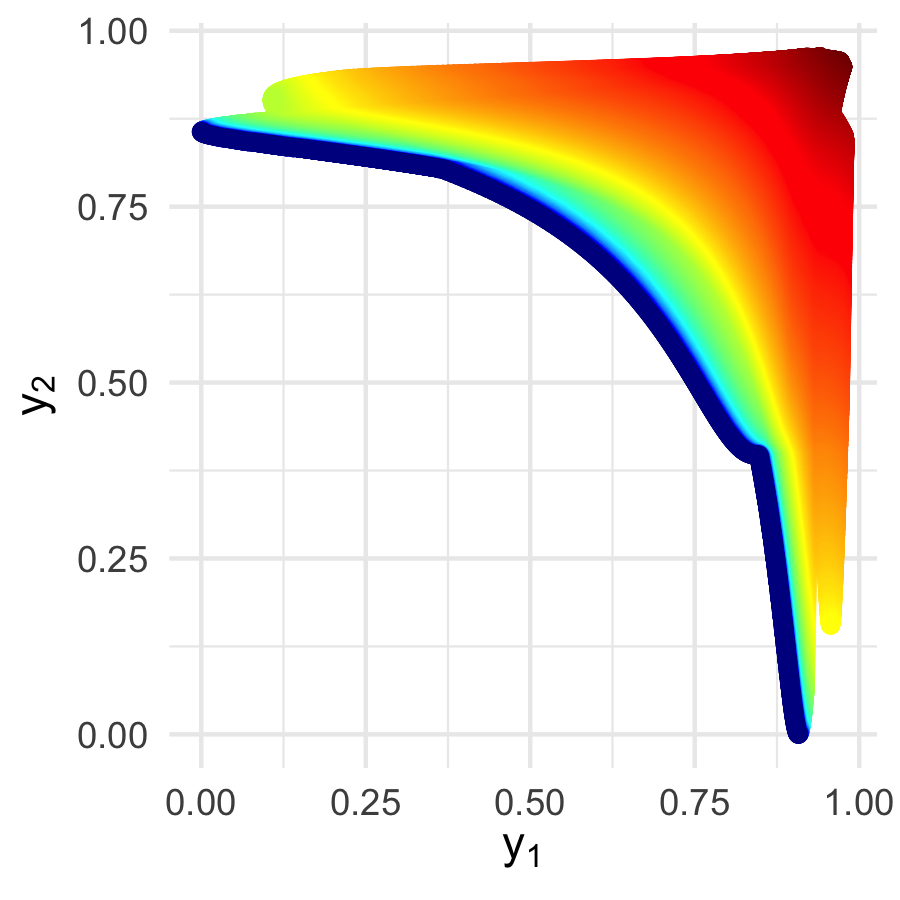}
    \hfill
    \includegraphics[width=0.32\textwidth]{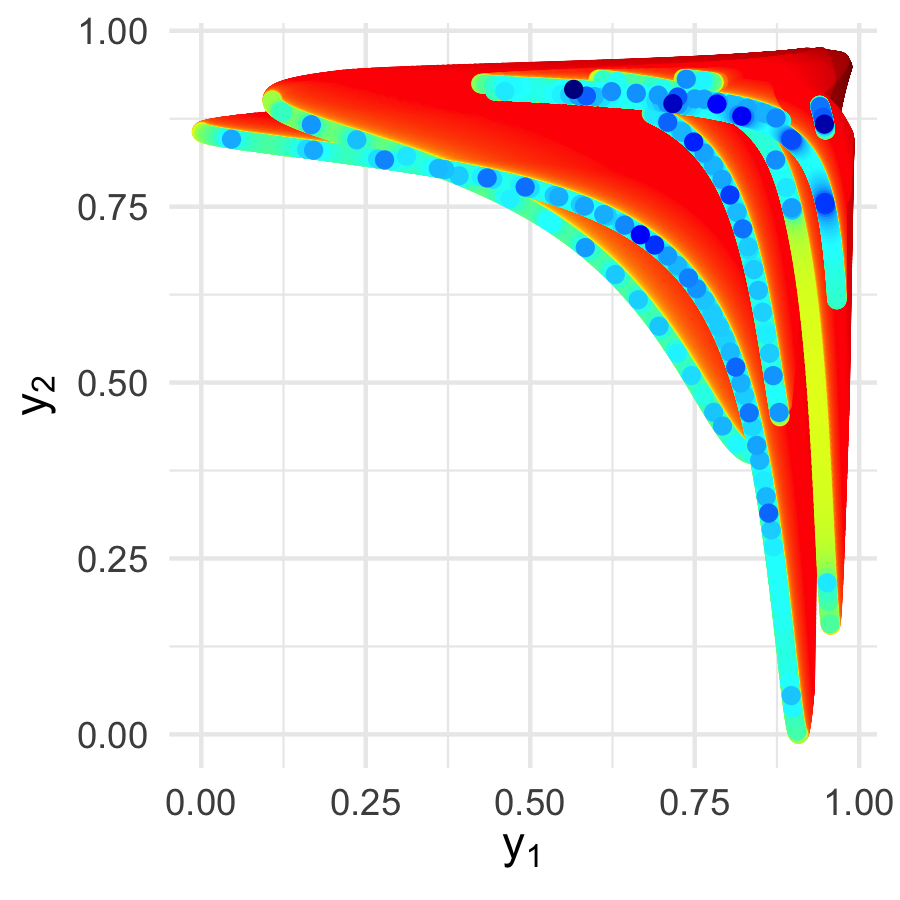}
    \hfill
    \includegraphics[width=0.32\textwidth]{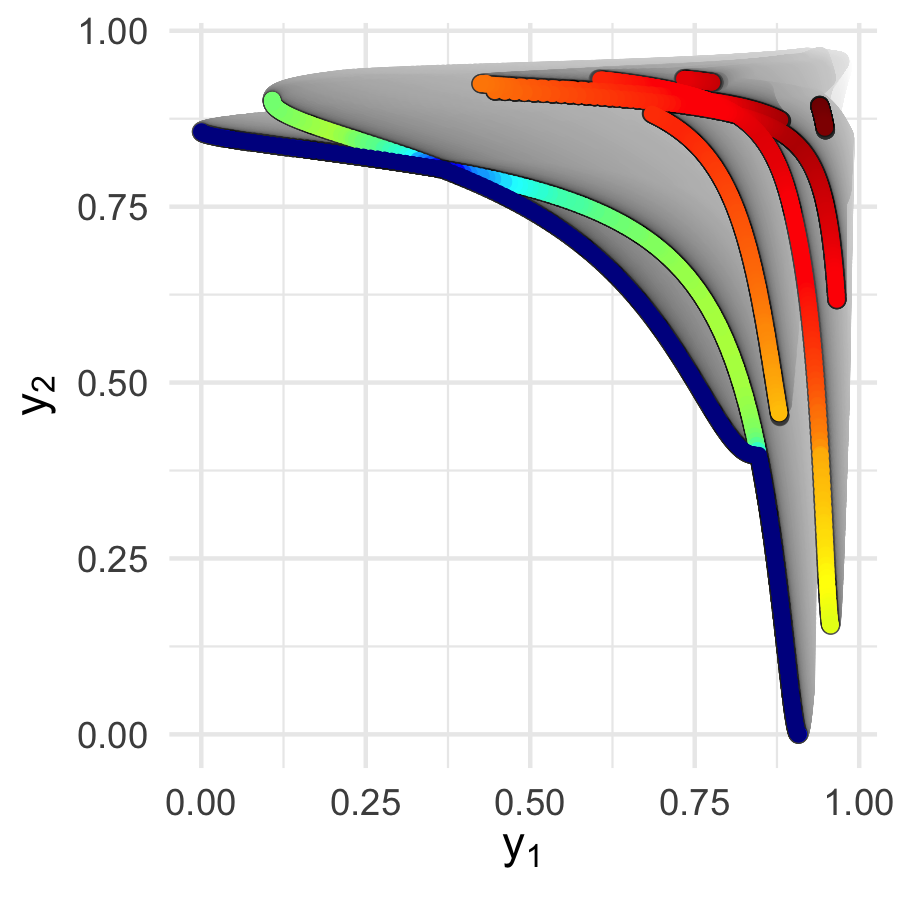}
    \vspace*{-0.4cm}
    \caption{Comparison of the cost landscape, gradient field heatmap, and PLOT visualizations (left to right), shown in the decision (top) and objective space (bottom). The color scales range from blue/black (better) to red/white (worse), respectively. The test function is a bi-objective MOP created with the MPM2 problem generator~\cite{wessing2015multiple} (with 2 decision variables, 3 peaks, random topology, and seeds 4 and 8, respectively). The visualizations reveal several locally optimal sets, which are frequently intersected by a superposed basin of attraction, and a disconnected globally optimal set.}
    \vspace*{-0.3cm}
    \label{fig:cost_gfh_plot}
\end{figure}

To our knowledge, there exist three main approaches for visualizing the interaction of decision and objective space of (two-dimensional) continuous MOPs. The cost landscape by Fonseca~\cite{Fonseca1995}, the gradient field heatmap by Kerschke and Grimme~\cite{kerschke2017expedition}, and the PLOT approach by this paper's authors~\cite{schapermeier2020plot}. Figure~\ref{fig:cost_gfh_plot} gives an overview of the different visualizations for an exemplary MOP.

A commonality of all three techniques is that they are based on a discretized version of the MOP's decision space. On the basis of this grid, each technique proceeds differently and focuses on particular properties of the landscape.

\medskip
\textbf{Cost Landscape~}
Being the first technique for illustrating the decision space of a MOP's landscape, Fonseca's \emph{cost landscapes} \cite{Fonseca1995} focus on \emph{global} optimality of points in the decision space -- but mainly ignore local relationships.
Therefore, the Pareto ranking is computed w.r.t. the evaluated points, and each point is assigned a height value corresponding to the number of points it is dominated by (plus one). These height values are then plotted on a logarithmic color scale, lending more weight to the (near) globally optimal points.

\medskip
\textbf{Gradient Field Heatmap~}
In contrast to the previous approach, the \emph{gradient field heatmap} by Kerschke and Grimme~\cite{kerschke2017expedition} gives insight into the (approximate) placements of \emph{locally} efficient sets and their corresponding basins of attraction.
This is achieved by computing the MOG (see section~\ref{sec:moo}) for each point in a discretized decision space, and subsequently moving to the neighbouring cell closest to this descent direction. This process is repeated until a locally efficient point is reached or an already seen cell is revisited. Along this integration path, the lengths of the MOGs are cumulated and interpreted as the ``height" value of the starting point. As with the cost landscape, the heights are then plotted on a logarithmic color scale. 

Note that a more recent, but less detailed visualization of local dominance and basins of attraction is provided in~\cite{fieldsend2019}. Therein, points from locally dominance-neutral regions or from basins of attraction are represented in binary fashion.

\medskip
\textbf{PLOT~}
The \emph{Plot of Landscapes with Optimal Trade-offs}~(PLOT)~\cite{schapermeier2020plot} is the most recent and most sophisticated of the visualization techniques, combining the advantages of the two previous approaches.
Based on the single- and multi-objective gradients, locally efficient points are approximated up to the resolution of the grid. Having these points determined explicitly, they can (1) be used to improve the visualization quality of the gradient field heatmap, and (2) be visualized in isolation from the remainder of the decision space. Within PLOT, the decision space is shown as a gray-scaled version of the gradient field heatmap, and all locally efficient points are colored according to their Pareto ranking. This provides complete information on the location(s) of the locally efficient points, their relation to global optima w.r.t. dominance, and their basins of attraction. Thereby it provides a comprehensive view of the structure of local and global optima for multi-objective benchmark problems.

\subsection{The Plotting Library \texttt{moPLOT}}

Although joint visualizations of the decision and objective space are essential to further our understanding of MOPs, so far, there exist only few works which considered these approaches. While one could argue that the gradient field heatmaps and PLOT visualizations are still rather new tools, the cost landscapes have been proposed two decades ago. Therefore, it is more likely that the limited usage of these approaches rather results from a lack of publicly available tools. We thus combined all the aforementioned visualization techniques within a single R-package called \texttt{moPLOT} (\url{https://github.com/kerschke/moPLOT}).

\section{The \texttt{moPLOT} Dashboard} \label{sec:dashboard}

As stated in the previous subsection, the R-package \texttt{moPLOT} provides all functionalities to produce the different visualizations. However, for users who do not have experience in using R, it will be very cumbersome to produce these images. To lower this burden and facilitate the usage of \texttt{moPLOT}, we therefore decided to enhance our plotting library with a graphical user interface (GUI).

When designing the interface, we wanted to satisfy a wide variety of requirements: it should (1) be accessible as a standalone and ideally web-based dashboard application that is independent of R, (2) contain all state-of-the-art methods for visualizing MOPs, (3) allow to produce plots for the decision and objective space, (4) support interactive usage (e.g., zooming into regions of interest), (5) offer easy access to a wide variety of common MOPs (from well-established test suites like DTLZ~\cite{Deb2005} or ZDT~\cite{Zitzler2000}, to more recent benchmarks like bi-objective BBOB~\cite{tusar2016} or the CEC 2019 test suite \cite{yue2019novel}), (6) enable the import of pre-computed data (to avoid repetitive time-consuming computations), and (7) allow to export the created visualizations to the user's hard disk.

In the following, we will briefly describe the chosen architecture and technology underlying our application, and afterwards give a short tour through the dashboard.

\subsection{Implementation}

We developed a user-friendly dashboard that meets all the desired requirements, available at \url{https://schaepermeier.shinyapps.io/moPLOT}.
At its core, the \texttt{moPLOT} dashboard is a web application based on the R-package \texttt{shiny}~\cite{chang2020shiny}. Plots are produced using the \texttt{ggplot2}~\cite{wickham2016} package and their interactivity is ensured by the graphing library \texttt{plotly}~\cite{sievert2020plotly}. In addition, we use the R-package \texttt{smoof}~\cite{bossek2017smoof} to easily access a plethora of benchmark functions.

\subsection{A Guided Tour Through the \texttt{moPLOT} Dashboard}

\begin{figure}[t!]
    \centering
    \includegraphics[width=\textwidth]{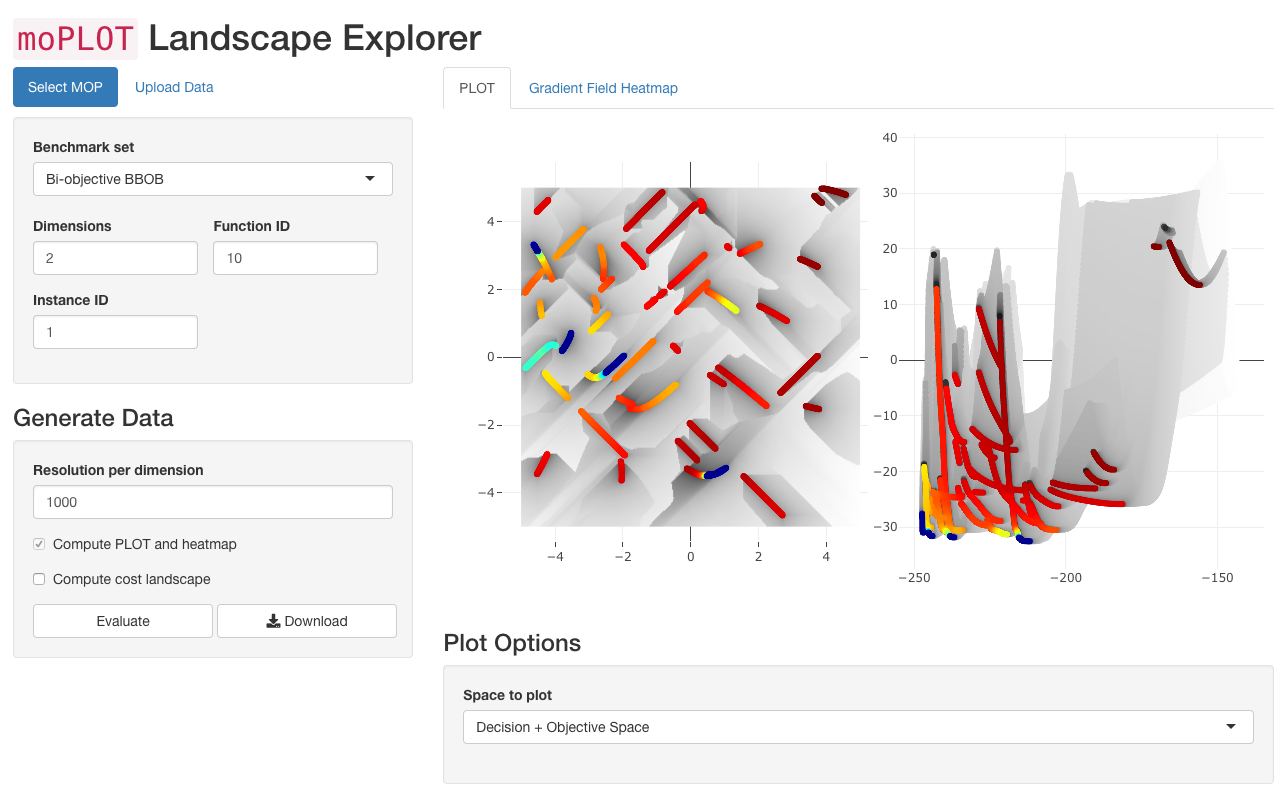}
    \vspace*{-0.8cm}
    \caption{Overview of the dashboard showing the PLOT visualization for a bi-objective BBOB function (Function ID: 10, Instance ID: 1) in the decision and objective space. In the left column, the user selects the test function, its parameters, and for which visualizations to generate data. Then, on the right, one of the available visualizations is selected, including some options for the chosen technique.}
    \vspace*{-0.3cm}
    \label{fig:dashboard_overview}
\end{figure}

Figure~\ref{fig:dashboard_overview} shows a screenshot of the dashboard's starting page. Initially, the user chooses a MOP (that is to be visualized) by first selecting the corresponding benchmark set (if applicable) and afterwards specifying the desired function. If a MOP is parametrizable, its parameters are made available in the interface, with default values already filled in. In the given example, we selected a function from the bi-objective BBOB \cite{tusar2016}.

After selecting the MOP, controls for the resolution of the visualizations and for generating the required visualization data will be revealed. They contain options on whether to compute the data for the PLOTs and gradient field heatmaps, or for the (computationally much more expensive) cost landscapes. Most of the time, these controls can be left at their corresponding default values. In the example, a grid resolution of $1,000 \times 1,000$ is used and data is generated only for the gradient field heatmap and PLOT approaches.

Subsequent to data generation, the visualizations are shown in the main view of the dashboard. Here, the user can choose between all visualizations for which the data is available. Separate views of the decision and objective space, as well as a joint view of both spaces, as shown in the example, are available.

\section{Visualization of Three-dimensional MOPs} \label{sec:three-dimensional}

Building on this new and interactive visualization environment, some of the previously introduced concepts can be enhanced to enable the visualization of three-dimensional MOPs. Extending the methods to handle three-dimensional decision spaces is beneficial for studying tri-objective MOPs in particular: While the Pareto sets of the latter are, in general, degenerated in two-dimensional decision spaces, they can be better studied in three-dimensional decision spaces.

In the following, we first discuss our extensions of the existing approaches for 3D decision spaces in section~\ref{sec:extend}. Afterwards, we introduce two techniques that enable the interactive exploration of an MOP's characteristics, inspired by MRI scans and isosurface-based ``onion layers'', respectively (section~\ref{sec:visualize3d}).

\subsection{Extension of Visualization Techniques}\label{sec:extend}

The amount of changes required to adapt each visualization technique to 3D decision and objective spaces varies greatly depending on the specific visualization.

\medskip
\textbf{Cost Landscape~}
For the cost landscape, virtually no changes need to be implemented. The Pareto ranking is not affected by a change in decision space dimensionality and naturally scales with the numbers of objectives. Yet, it may become more computationally expensive. The only issue is the visualization of a 3D volume of data points (called \emph{voxels} in graphics research), which will be addressed in the following subsection.

\medskip
\textbf{Gradient Field Heatmap~}
The gradient field heatmap requires some changes to handle 3D decision and objective spaces, respectively.
To cover a 3D decision space, the approach used for following the gradient from one cell to the next one is changed such that decisions are made per decision space dimension. More precisely, for each variable $x_i$, $i = 1, 2, 3$, we compute the angle between the MOG and the plane spanned by $\{x_1, x_2, x_3\} \setminus \{x_i\}$. If that angle exceeds $22.5^{\circ}$ (is below $-22.5^{\circ}$), a step will be made in positive (negative) direction of $x_i$, otherwise one will keep the $x_i$ value of the current cell. This decision is made for all $p = 3$ dimensions and by combining all results, we identify the next cell.

Moving from two to three objectives requires an additional change. Originally, the gradient field heatmap was only defined for bi-objective problems using the sum of normalized single-objective gradients as MOG. As introduced in section \ref{sec:moo}, the MOG based on the convex hull can be used to generalize the MOG for more objectives. With the updated MOG definition, the remainder of the approach can remain unchanged.

\medskip
\textbf{PLOT~}
As PLOT builds on the other two visualizations, some of the main extensions have already been discussed. The only further component requiring changes is the approximation of the locally efficient points in a 3D decision space.

For a 2D PLOT, first-order critical points are determined based on the convex hull of the single-objective gradients in triangular neighborhoods. If the zero vector is contained in its interior, the points in this neighborhood are considered first-order critical. This directly extends to 3D using tetrahedral neighbourhoods.

Then, a stability analysis of the MOG vector field (oriented towards the descent direction) is conducted to ensure second-order optimality. It is considered stable, if the eigenvalues of its Jacobi matrix all have non-positive real parts. In 2D, this condition can be checked robustly using the MOG's divergence \cite{schapermeier2020plot}. Analogously, for 3D, robust conditions based on matrix invariants are available~\cite{chong1990general}. Note that the critical points are degenerate (i.e. not singular) w.r.t. the MOG vector field, such that we can expect at least one eigenvalue to be zero.

\subsection{Visualizing a 3D Volume of Points}\label{sec:visualize3d}

Visualization of 3D volume elements (i.e., the voxels or regular cells, which fill out the considered 3D decision space) comes with the challenge of outer voxels obscuring the view from inner voxels. For overcoming this issue, we propose to use two different approaches: An MRI scan-like slicing of the volume~\cite{hansen_visualization_2005} and an onion-like layer-based visualization based on isosurfaces~\cite{udupa_display_1983}.

\medskip
\textbf{MRI Scan~}
\begin{figure}[t!]
    \centering
    \includegraphics[width=.49\textwidth, trim = 0mm 0mm 0mm 23mm, clip]{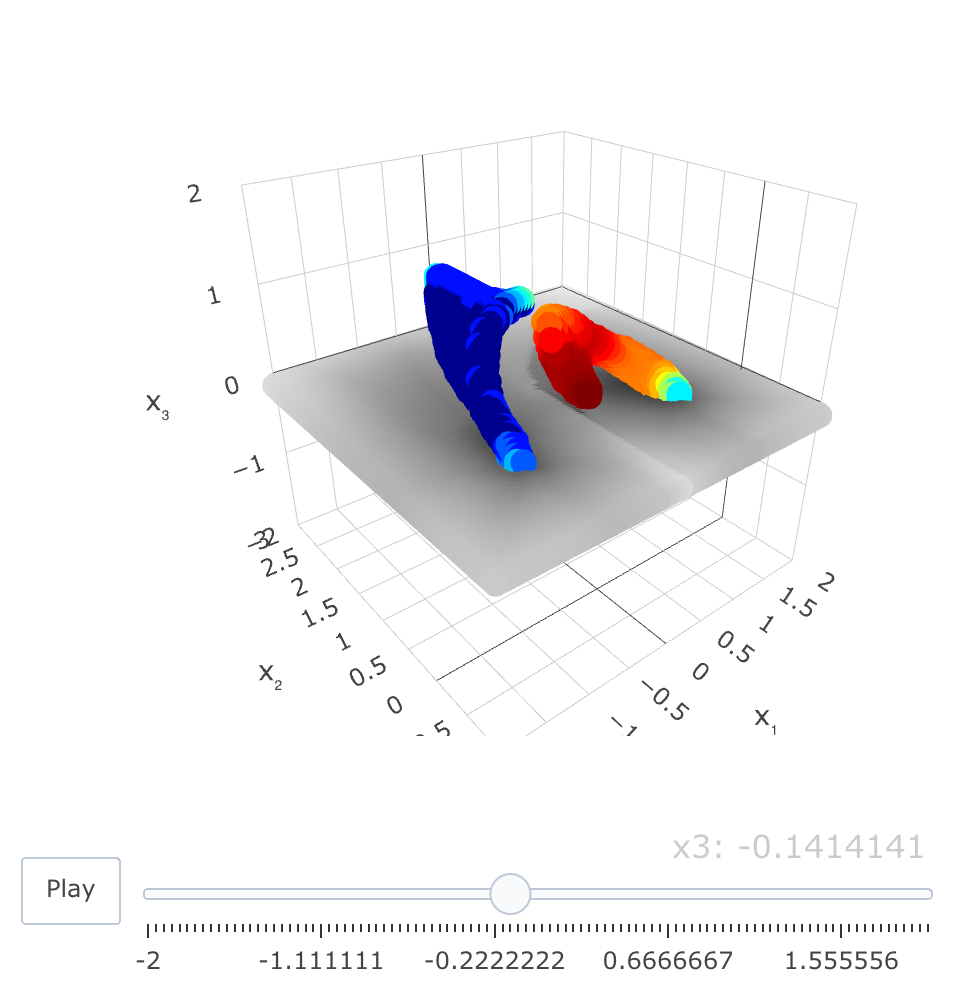}
    \hfill
    \includegraphics[width=.49\textwidth, trim = 0mm 0mm 0mm 23mm, clip]{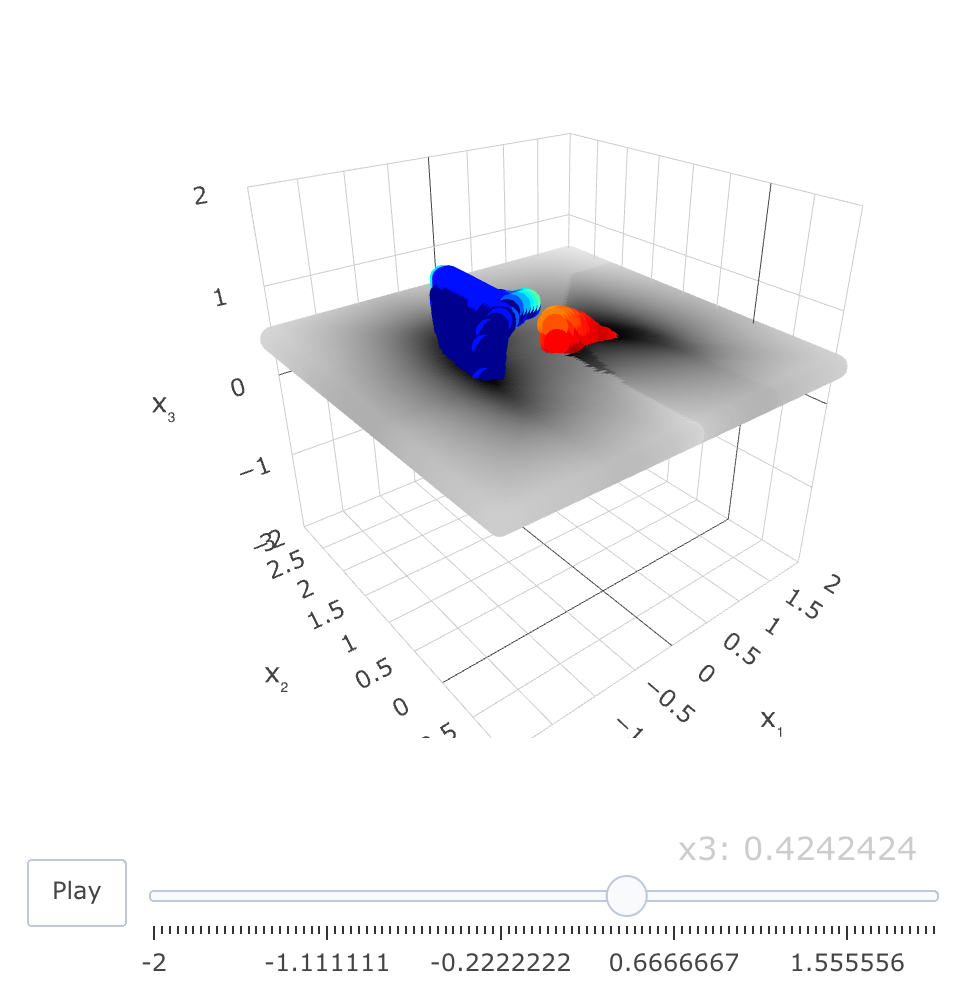}
    \vspace*{-0.2cm}
    \caption{The MRI scan technique applied to the PLOT visualization for a tri-objective MOP with one locally optimal (red) and one globally optimal (blue) set in the decision space. The user can slide the heatmap plane along the $x_3$ direction (another dimension may also be chosen), showing the intersection of the attraction basins of the two sets.}
    \vspace*{-0.2cm}
    \label{fig:mri_scan_plot}
\end{figure}
A general technique to reduce the amount of data shown is by \textit{slicing} it along one of the three coordinate axes. Then, the placement along this axis can be controlled by the user, providing a complete picture of the visualization data. As this resembles an MRI scan, we use this name to refer to this method.

In the case of the PLOT technique, as there are only few locally efficient points, these can be permanently visualized, applying the slicing only to the gradient field heatmap. An example of this is given in figure \ref{fig:mri_scan_plot}.

\medskip
\textbf{Onion Layers~}
\begin{figure}[t!]
    \centering
    \includegraphics[width=.49\textwidth, trim = 0mm 0mm 0mm 8mm, clip]{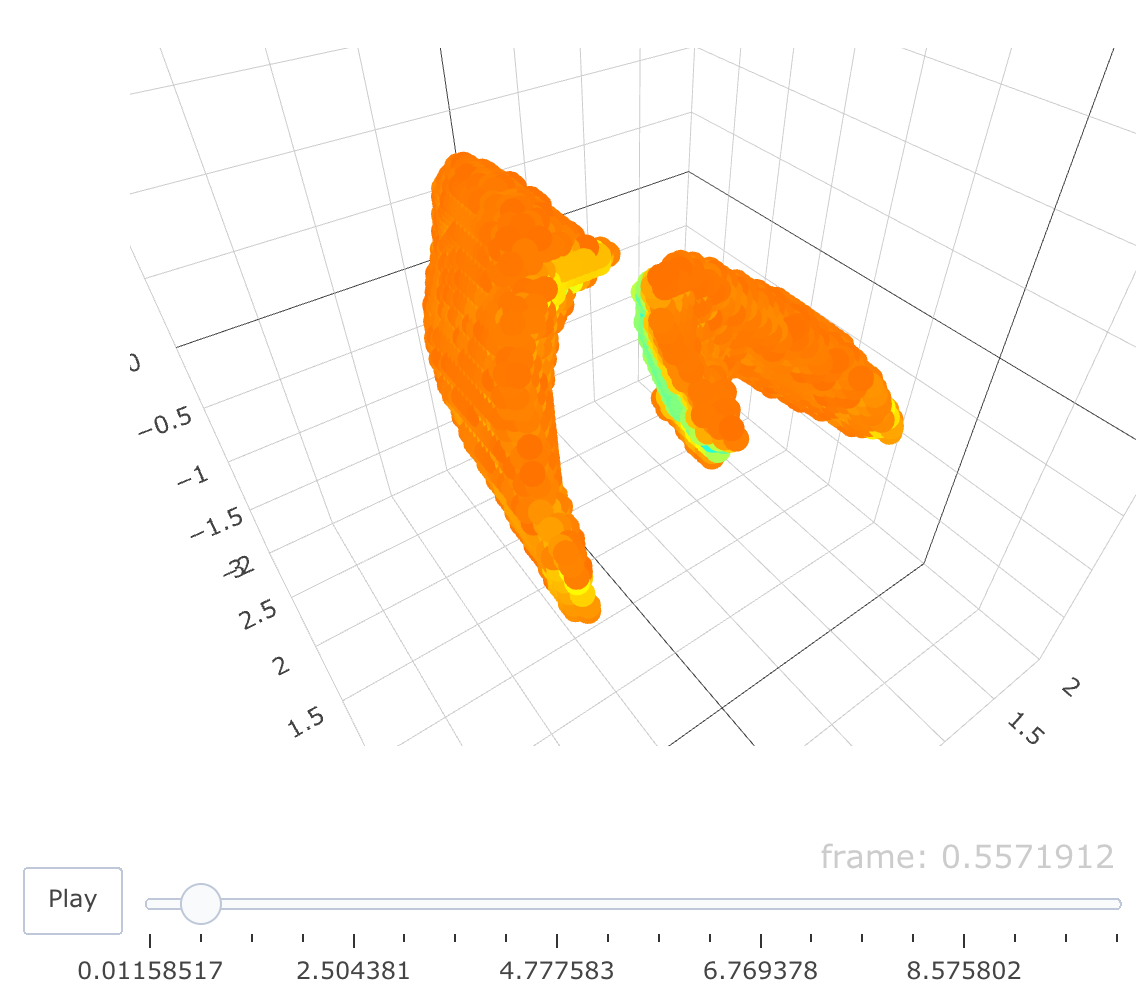}
    \hfill
    \includegraphics[width=.49\textwidth, trim = 0mm 0mm 0mm 8mm, clip]{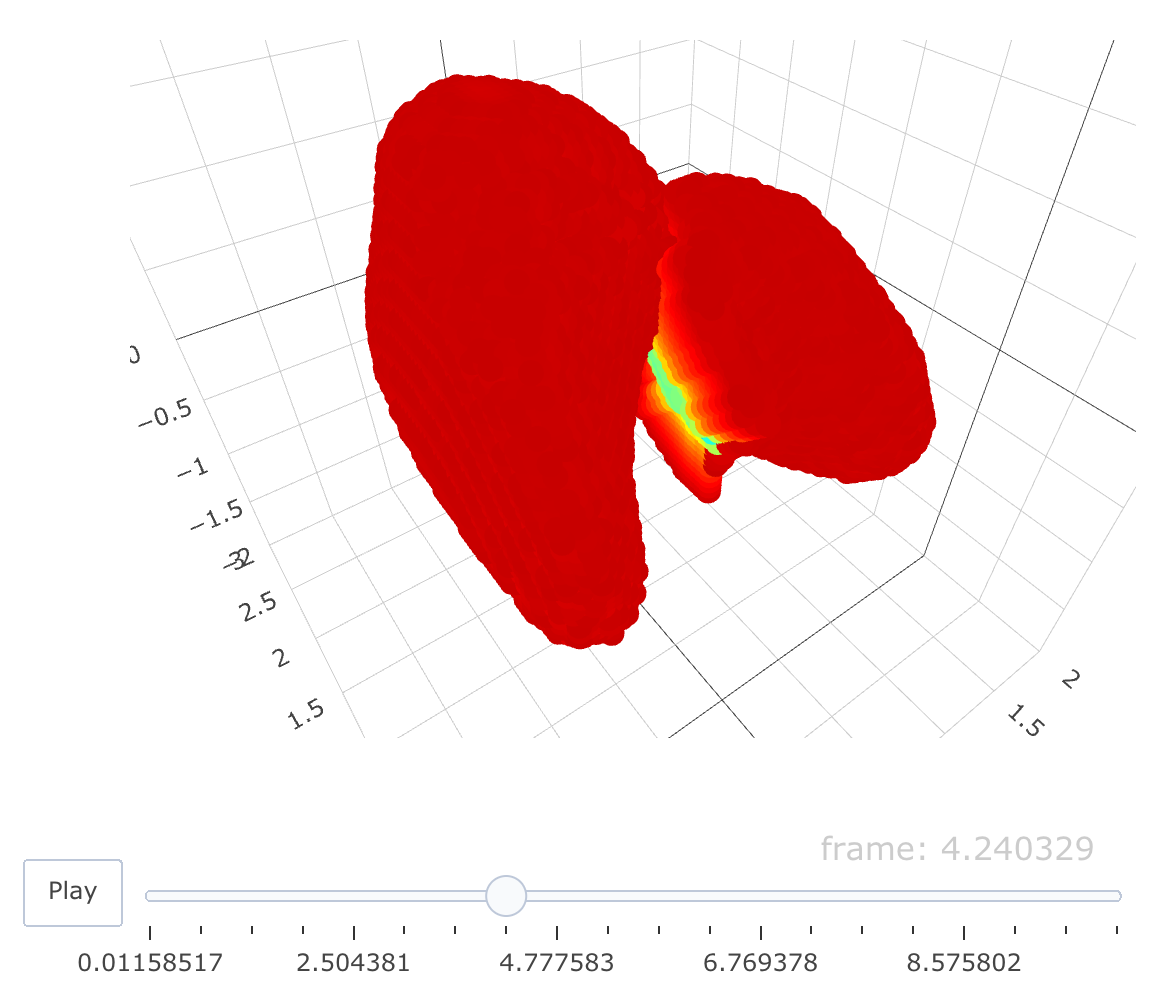}
    \vspace*{-0.2cm}
    \caption{The onion layers technique applied to the gradient field heatmap visualization for the same tri-objective MOP as in figure \ref{fig:mri_scan_plot}. The value of the isosurface can be adjusted by the user. Again, it can be observed that the basin of the locally optimal set is superposed by the basin of the globally optimal set.}
    \vspace*{-0.4cm}
    \label{fig:onion_layers_heatmap}
\end{figure}
Based on the gradient field heatmap, we propose an alternative view on attraction basins that literally encapsulate the (locally) efficient sets in 3D decision space. Therefore, we consider the discretized isosurface\footnote{A level set of a function $f:\mathbb{R}^n\to\mathbb{R}$ w.r.t. value $c\in\mathbb{R}$ is the set of points $\{(x_1,\dots,x_n)\in\mathbb{R}^n:f(x_1,\dots,x_n) =c\}$. The isosurface is a 3D level set. } w.r.t. the computed voxels at a given height value. Different height thresholds reveal layers of the discretized decision space and allow the analysis of the volume and the attraction basins. As we could interpret the different isosurfaces as layers surrounding the efficient sets, we name this approach onion layer visualization.

In addition to illustrating the approximate location of the locally efficient points, the \textit{onion layers} also reveal the intersection of their corresponding attraction basins. An example of this method is given in figure \ref{fig:onion_layers_heatmap}.

\section{Conclusions} \label{sec:conclusions}

In this paper, we introduced the R-package \texttt{moPLOT} and its accompanying web-based, interactive and user-friendly dashboard. Thereby, we lowered the technical barrier to access and interactively explore the state-of-the-art techniques for visualizing continuous multi-objective optimization problems (MOPs). Building on this interactive environment, we additionally extended the existing visualization techniques to also enable illustrations of 3D decision and objective spaces. This extension leads to a substantial expansion of the range of visualizable MOPs.

The \texttt{moPLOT} package and dashboard can be used as a basis for further visualization techniques. For example, the existing, comprehensive visualizations do not scale beyond a 3D decision space due to a lack of intuitively visualizable dimensions. An avenue for further research is presented by applying different techniques in reducing this dimensionality. For example, one could explore the interactions between different local efficient sets (such as superposed basins), akin to local optima networks \cite{ochoa2008study}. Dimensionality reduction techniques based on (random) cut planes, as applied, e.g., in visualizing the loss landscapes of neural networks \cite{li2018visualizing} or (bi-objective) BBOB functions \cite{hansen2009,tusar2016}, may also give insights into the decision space of higher-dimensional MOPs.

Beyond gaining insights on the optimization landscapes themselves, another future extension can cover the visualization of optimizer trajectories. This would allow to study the interactions between landscape properties and the performance of different optimization algorithms.

\section*{Acknowledgements}
The authors acknowledge support by the \href{https://www.ercis.org}{\textit{European Research Center for Information Systems (ERCIS)}}.

\bibliographystyle{unsrt}
\bibliography{arxiv}

\end{document}